\documentclass[conference]{IEEEtran}
\IEEEoverridecommandlockouts
\usepackage{cite,amsmath,amssymb,graphicx,booktabs,array}
\usepackage[hidelinks]{hyperref}
\hypersetup{pdftitle={Structure vs. Chain-of-Thought: Evaluating LLM Criteria Extraction for Depression Severity},
  pdfauthor={Xinkai Chen}}
\usepackage{etoolbox}
\usepackage[T1]{fontenc}
\makeatletter
\patchcmd{\@makecaption}{\scshape}{}{}{\errmessage{table caption patch failed}}
\makeatother

\def\BibTeX{{\rm B\kern-.05em{\sc i\kern-.025em b}\kern-.08em T\kern-.1667em\lower.7ex\hbox{E}\kern-.125emX}}
\newcommand{\sig}[1]{\textbf{#1}}
\newcommand{\appref}[1]{\hyperref[#1]{Appendix~\ref*{#1}}}

\begin{document}

\title{Structure vs. Chain-of-Thought: Evaluating LLM\\
Criteria Extraction for Depression Severity%
\thanks{Extended version of a paper accepted at the 2nd Workshop on Mental Health Disorder
Detection on Social Media (MHSM 2026), held with IEEE ICDM 2026. It restores material cut
from the six-page proceedings version and adds \hyperref[app:comp]{Appendices~\ref*{app:comp}}--\hyperref[app:prompts]{\ref*{app:prompts}};
numbers that appear in both versions are identical.}}

\author{\IEEEauthorblockN{Xinkai Chen}
\IEEEauthorblockA{Independent Researcher\\
\href{mailto:xinkaichen1997@gmail.com}{xinkaichen1997@gmail.com}}}

\maketitle

\begin{abstract}
A large language model (LLM) can rate depression severity directly from a social media post,
or mark which clinical criteria the post shows and let a simple rule turn the count into a label.
The latter is easier to audit because a clinician can check each marked criterion.
We compare these approaches on two Reddit corpora using three LLMs (from 9B to frontier
scale) and two questionnaires (PHQ-9, BDI-II), and measure agreement with quadratic
weighted kappa ($\kappa_w$). For the two frontier models, criteria extraction scores above
chain-of-thought on one corpus only when its decision thresholds are fitted on labeled data.
Neither model's gain is significant, with or without recalibrating chain-of-thought on the same labels.
With thresholds fixed \emph{a priori} from PHQ-9's criteria, extraction shows no gain on
either corpus, even where models mark over two criteria per post. The 9B model behaves
differently on a corpus from depression communities. It labels most posts as severe, whether prompted
directly or with chain-of-thought, while the \emph{a priori} rule, which uses no labels, beats both. After
chain-of-thought is recalibrated on the same labels, no significant gap remains, consistent
with a calibration effect.
Yet higher ordinal agreement does not ensure better detection of severe cases. PHQ-9
criteria extraction misses most severe posts, and moving from direct prompting to
chain-of-thought and then to extraction increases misses in nearly all comparisons.
On one corpus, a relabeled stress dataset, a model using that dataset's own features,
some computed from the post text, is not significantly different from frontier criteria
extraction under the \emph{a priori} rule.
\end{abstract}


\section{Introduction}

Depression severity in social media posts is usually predicted as an ordinal label, often
with large language models (LLMs). One appealing design is to split the task. DSM-5, the standard
psychiatric manual, lists nine symptoms of a major depressive episode (its criterion~A), and
the PHQ-9 questionnaire~\cite{kroenke2001} asks about exactly these nine. An LLM marks which
criteria a post shows, and a simple rule turns the count into a label that a clinician can check.
This design has one practical advantage: it is far more stable than
chain-of-thought across identical runs, so it is easier to audit.

We test this design on two Reddit corpora, DepSeverity~\cite{naseem2022} (four severity
labels) and DepSign~\cite{sampath2022,kayalvizhi2022} (three labels), and measure agreement with the gold labels using
quadratic weighted kappa ($\kappa_w$)~\cite{cohen1968}, which is 0 at random chance and weights each error by its squared distance from the true label. We test three models: a small local one (Qwen3.5-9B), an
open-weights frontier model (DeepSeek-V4.1-Flash) and a closed one (Claude-Sonnet-5).

We make four contributions.

\textbf{(1) An unfair comparison, measured.} On DepSeverity, criteria extraction with fitted
thresholds scores above chain-of-thought for all three models. With thresholds fixed
\emph{a priori}, all three gains reverse sign (at worst $-0.081$). A pipeline that fits thresholds on labels should
therefore not be compared directly with a zero-shot prompt that sees no labels: the difference mostly
shows the value of the labels, not of the pipeline.

\textbf{(2) A weak model behaves differently.} For the 9B model, criteria extraction with fitted thresholds leads
chain-of-thought by a wide margin: $+0.149\,\kappa_w$ on DepSeverity, against $+0.022$ and
$+0.037$ for the frontier models (neither significant), and $+0.109$ remains when chain-of-thought
gets the same labels (significant before correction). With \emph{a priori} thresholds the gain
disappears ($-0.018$). On DepSign, the 9B model labels most posts as \textsc{severe}, directly and with chain-of-thought,
and the \emph{a priori} rule beats both ($+0.093$ and $+0.143$, respectively);
counting BDI-II's 21 items instead, it still beats chain-of-thought ($+0.108$).
This is consistent with a calibration effect: the fixed rule reduces the model's
over-prediction of \textsc{severe}, and recalibrating chain-of-thought with the same 600 labels
leaves no significant gap. With BDI-II's items, the 9B model also beats chain-of-thought without
labels on DepSeverity ($+0.130$), though not significantly once chain-of-thought is
recalibrated ($+0.091$).

\textbf{(3) Better ordinal agreement, more missed severe posts.} From direct prompting to
chain-of-thought, and from chain-of-thought to criteria extraction, the share of
gold-\textsc{severe} posts predicted below \textsc{severe} (false negatives) rises in 11 of 12 comparisons
(three models, two corpora), significantly in 8 before correction. \textsc{severe} is rare, 7--8\% of posts, so an approach that rarely
predicts it scores higher on $\kappa_w$ and catches fewer \textsc{severe} posts. The approach with the lowest
$\kappa_w$ catches that class best, and criteria extraction's gain on the 9B model comes with
47 of 50 \textsc{severe} posts missed. An ablation that keeps the criteria but lets the model
assign the label shows where the loss arises: counting the criteria, not reasoning about them.

\textbf{(4) Much of one benchmark can be predicted from its source dataset's own fields.}
DepSeverity is a relabeled copy of Dreaddit~\cite{turcan2019}, a stress dataset, and none of its source
communities is about depression. As a check on the benchmark itself, a model that uses only
Dreaddit's own features (including source community, stress label, and word-category counts computed from
the text) reaches $\kappa_w = 0.404$, not significantly different from frontier criteria
extraction with \emph{a priori} thresholds. Dreaddit's fields that do not read the post reach 0.333,
above every zero-shot direct prompt.

All comparisons use paired bootstrap tests on the same posts.

\section{Related Work}

Prior work annotates posts criterion by criterion rather than with a single severity label:
the Depressive Disorder Annotation scheme~\cite{mowery2015} covers the DSM-5 criteria, and PRIMATE~\cite{gupta2022} labels PHQ-9
criteria in Reddit posts and D2S~\cite{yadav2020} in tweets, although a mental-health professional
who re-annotated PRIMATE found many false positives for anhedonia~\cite{milintsevich2024}. We instead evaluate
extraction end to end, against severity labels, and study the step that turns criteria into
a label. The eRisk tasks fill in BDI-II from a user's whole posting history~\cite{erisk2019}
and rank sentences by relevance to each of its 21 symptoms~\cite{erisk2023}; we instead mark
the 21 items as present or absent in one post and count them (C4,
\S\ref{sec:conditions}).

Fitting thresholds is fair when all compared systems use the same labels. Next to a zero-shot prompt, however, fitted thresholds
confound the comparison, and this setup is common: supervised classifiers on LLM embeddings beat zero-shot prompting on
severity, leading to the view that LLMs work better as interpreters than as
classifiers~\cite{kim2026}, and guidelines learned from labeled examples beat zero-shot
prompting on BDI-II items~\cite{bao2026}. We know of no work that measures this effect.
Chain-of-thought prompting~\cite{wei2022} is our control: the model reasons first but still
gives the label itself. On DepSeverity, Cognitive-Mental-LLM~\cite{patil2025} finds
chain-of-thought \emph{below} direct prompting in accuracy, the opposite of our result; our
direct prompts over-predict severity, which chain-of-thought corrects (\S\ref{sec:control}),
so the direction may depend on how well calibrated the direct prompt is.

\section{Corpora}

\subsection{DepSeverity and its origin}\label{sec:prov}

DepSeverity~\cite{naseem2022} contains 3{,}553 Reddit posts labeled \textsc{minimum},
\textsc{mild}, \textsc{moderate} or \textsc{severe}, with no stated license.

DepSeverity reuses Dreaddit's posts, as noted by Mental-LLM~\cite{xu2024}, which evaluates on
both, and by a later study~\cite{ibrahimov2025}. We verify it: after normalizing whitespace, \textbf{all 3{,}530
de-duplicated posts appear word for word in Dreaddit}, so DepSeverity is a relabeling, not an
extension. A model trained or tuned on Dreaddit has therefore already seen every DepSeverity
test post, under a different label, and its DepSeverity results may reflect that exposure. Dreaddit also records each post's subreddit, which DepSeverity drops and, to our
knowledge, no earlier work has recovered. The
communities are \textit{ptsd}, \textit{relationships}, \textit{anxiety},
\textit{domesticviolence}, \textit{assistance}, \textit{survivorsofabuse},
\textit{homeless}, \textit{almosthomeless}, \textit{stress} and \textit{food\_pantry}. \textbf{None is a
depression community.} Accordingly, only 3 of 56 gold-\textsc{severe} test posts mention self-harm.

Dreaddit's own split is stratified by stress and leaves only 10 \textsc{severe} and 9
\textsc{mild} test posts, so we make our own stratified 80/20 split with a fixed seed. The annotation documents are not available, so
we separate what is known from what we infer. The four label names are the severity bands of
the BDI-II questionnaire~\cite{beck1996} (PHQ-9 has five), and~\cite{naseem2022} reports
annotating with Beck's questionnaire and the Depressive Disorder Annotation
scheme~\cite{mowery2015}. We infer that only the band names come from BDI-II, while the
symptom-level scheme uses the nine DSM-5 criteria that PHQ-9 asks about. If that inference is wrong, and the annotators judged severity with BDI-II's
21 items, PHQ-9 is not the right instrument; condition C4
(\S\ref{sec:conditions}) therefore repeats the extraction over those 21 items.

We drop 4 posts in the two duplicate groups whose copies carry different gold labels (one
\textsc{minimum}/\textsc{severe} pair on identical text) and merge 19 other duplicates that
would otherwise span train and test, leaving 3{,}530 posts.

\subsection{DepSign}

DepSign~\cite{sampath2022} comes from mental-health subreddits, including
\textit{r/depression} and \textit{r/MentalHealth}, is labeled \textsc{not depression},
\textsc{moderate} or \textsc{severe}, and has no stated license. Two domain experts labeled the posts against written
guidelines, with Cohen's $\kappa = 0.686$ between them~\cite{sampath2022}. We subsample
its \emph{official} splits to DepSeverity's sizes (706 test, 600 train;
Table~\ref{tab:corpora}), so the corpora differ in class balance (73\% \textsc{minimum}
against 67\% \textsc{moderate}) but not in test size.

\textbf{1{,}502 of DepSign's 16{,}632 posts (9.0\%) contain a} \texttt{[removed]} \textbf{or}
\texttt{[deleted]} \textbf{tag} and have a median of 11 words, but were still labeled: 57\%
as \textsc{not depression}, against 25\% of the other posts, so a model that learns ``little
text $\rightarrow$ not depression'' gains accuracy for free. DepSeverity has no such posts. We
keep them, so the split stays official, and test their effect in \S\ref{sec:empty}.

\begin{table}[!tb]
\caption{Test splits. Both corpora are evaluated at $n{=}706$.}
\label{tab:corpora}
\centering\small
\setlength{\tabcolsep}{4pt}
\begin{tabular}{lcc}
\toprule
 & \textbf{DepSeverity} & \textbf{DepSign} \\
\midrule
lowest class & 513 \textsc{minimum} & 184 \textsc{not dep.} \\
middle class(es) & 58 \textsc{mild} / 79 \textsc{mod.} & 472 \textsc{mod.} \\
highest class & 56 \textsc{severe} & 50 \textsc{severe} \\
\midrule
majority acc. & 0.727 & 0.669 \\
majority $\kappa_w$ & 0.000 & 0.000 \\
median words & 80 & 104 \\
\bottomrule
\end{tabular}
\end{table}

\section{Method}

\subsection{Conditions}\label{sec:conditions}

\textbf{C1 (direct).} The model returns only a severity label.

\textbf{C2 (chain-of-thought).} The model first reasons about which depressive symptoms the
post shows, then gives the label itself.

\textbf{C3 (structured, PHQ-9).} The model returns JSON that marks each of the nine PHQ-9
criteria as \texttt{present}, \texttt{absent} or \texttt{unclear} (kept separate from
\texttt{absent}), with a word-for-word quote from the post for each \texttt{present}.
\textbf{The model never sees the severity labels and never outputs one}; a counting rule
assigns the label.

\textbf{C4 (structured, BDI-II).} Repeats C3 over the 21 items of BDI-II, the questionnaire whose band
names DepSeverity uses; these items only partly overlap with the DSM-5 criteria.

\textbf{C2S (ablation: criteria-structured chain-of-thought).} As C2, but the model first
writes one line per PHQ-9 criterion, giving its status (\texttt{present}, \texttt{absent} or
\texttt{unclear}) and a brief reason, and then gives the label itself. C2 and C3 differ in
three ways at once: what the model reasons about, the output format, and who assigns the label.
C2S keeps C2's format and labeler but reasons over C3's criteria, and its status lines can also
be counted with C3's rule (``C2S counted''), which changes only the labeler.

\subsection{From criteria to a label}\label{sec:agg}

Let $s$ be the number of criteria (C3) or items (C4) marked \texttt{present}. We turn $s$
into one of $k$ ordinal labels with $k-1$ thresholds, set in two ways.

\textbf{\emph{Fitted.}} We search all threshold settings and keep the one with the highest
$\kappa_w$ on a 600-post fitting split taken only from training data. We then freeze these thresholds
for the test set.

\begin{figure}[!tb]
\centering
\includegraphics[width=\columnwidth]{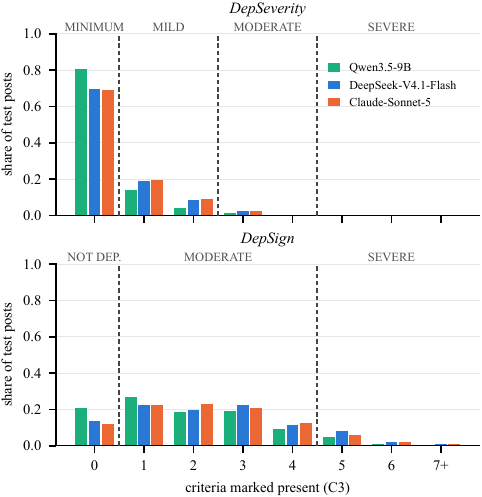}
\caption{Criteria marked \texttt{present} per test post under C3, by model, with the
\emph{a priori} thresholds and the labels they assign.}
\label{fig:counts}
\end{figure}

\textbf{\emph{A priori.}} The thresholds are label-free: we fix them without looking at any label.
For the nine PHQ-9 criteria, DSM-5 requires at least five symptoms for a major depressive
episode (one must be depressed mood or loss of interest; every post reaching five here has one), so the
top threshold is $4.5$. A post with no symptom gets the lowest label, so the first threshold
is $0.5$. DSM-5 says nothing about counts of $1$--$4$: we split them evenly for DepSeverity's four labels,
$[0.5, 2.5, 4.5]$, and add no middle threshold for DepSign's three, $[0.5, 4.5]$. The 21 BDI-II
items have no exact anchor, so we report C4 with fitted thresholds and two imperfect
\emph{a priori} rules: the DSM-5 thresholds above, and BDI-II's bands rescaled to 0--21.
On DepSeverity the models mark no criterion in most posts, so the two upper labels are rarely reached;
on DepSign the counts spread across the scale (Fig.~\ref{fig:counts}).

\textbf{Stricter thresholds do not help.} We also try requiring seven symptoms for \textsc{severe}
($[1.5, 4.5, 6.5]$, our choice, not a DSM-5 rule). On DepSeverity this lowers $\kappa_w$ to
0.231 and 0.223 for the frontier models and to 0.117 for the 9B model. On DepSign, depending on
how the four bands map onto three labels, it never significantly beats C2 for the frontier
models ($\kappa_w$ 0.04--0.28); for the 9B model its gain over C2 is at most $+0.092$, below the
standard rule's $+0.143$. It catches no gold-\textsc{severe} post on either corpus: such cut-offs suit a clinical interview, which asks about every
symptom, while a post mentions only a few.

The difference between the two regimes is central: fitted thresholds use labels, so C3 is no
longer zero-shot while C1 and C2 still are, which confounds a direct comparison.

The score counts only \texttt{present}; C3 and C4 mark a criterion \texttt{absent} in only
0.06--0.54\% of judgments, so uncounted criteria are almost all \texttt{unclear}.

\subsection{Models and protocol}

We use Qwen3.5-9B (run locally, open weights), DeepSeek-V4.1-Flash (open weights) and
Claude-Sonnet-5, with extended thinking turned off, because hidden reasoning in C1 would make
the C1/C2 comparison meaningless. We verified this for each vendor and count reasoning leaks on every call. Temperature is 0 where
the API allows it, though no hosted model is deterministic even then, and Claude-Sonnet-5
accepts no sampling parameters at all. We therefore report bootstrap intervals over test posts,
\textbf{paired} across conditions, and measure run-to-run variation in \S\ref{sec:stab}. We cache and log every response.

\section{Results}

\begin{table}[!tb]
\caption{Test-set results, $n{=}706$ per corpus. $\kappa_w$ = quadratic weighted kappa; F1$_M$ = macro-F1;
\textsc{sev}$\downarrow$ = fraction of \textsc{severe} posts given a lower label ($1-{}$recall on
that class). C3 rows show
fitted~/~\emph{a priori} (DSM-5) thresholds, C4 rows fitted~/~DSM-5~/~BDI-II bands; other
columns are fitted. Bold: highest $\kappa_w$ per model and corpus, fewest missed
\textsc{severe} posts per corpus.}
\label{tab:main}
\centering\small
\setlength{\tabcolsep}{2.5pt}
\begin{tabular}{llccccc}
\toprule
\textbf{Model} & \textbf{Cond.} & $\kappa_w$ & MAE & Acc & F1$_M$ & \textsc{sev}$\downarrow$ \\
\midrule
\multicolumn{7}{l}{\emph{DepSeverity} (4 classes, 56 \textsc{severe})}\\
\midrule
Qwen3.5-9B   & C1 & 0.288 & 0.948 & 0.445 & 0.335 & 29/56 \\
             & C2 & 0.340 & 0.877 & 0.456 & 0.327 & 34/56 \\
             & C3 & 0.489~/~0.322 & 0.465 & 0.728 & 0.367 & 43/56 \\
             & C4 & \sig{0.495}/0.470/0.057 & 0.517 & 0.643 & 0.420 & 40/56 \\
DeepSeek     & C1 & 0.219 & 1.280 & 0.329 & 0.251 & \sig{16/56} \\
V4.1-Flash   & C2 & 0.504 & 0.487 & 0.659 & 0.408 & 42/56 \\
             & C3 & \sig{0.526}~/~0.423 & 0.452 & 0.670 & 0.362 & 49/56 \\
             & C4 & 0.469/0.479/0.079 & 0.589 & 0.592 & 0.386 & 41/56 \\
Claude       & C1 & 0.299 & 0.963 & 0.354 & 0.306 & 33/56 \\
Sonnet-5     & C2 & 0.462 & 0.551 & 0.596 & 0.390 & 41/56 \\
             & C3 & \sig{0.499}~/~0.395 & 0.467 & 0.664 & 0.369 & 49/56 \\
             & C4 & 0.460/0.466/0.110 & 0.616 & 0.508 & 0.347 & 48/56 \\
\midrule
\multicolumn{7}{l}{\emph{DepSign} (3 classes, 50 \textsc{severe})}\\
\midrule
Qwen3.5-9B   & C1 & 0.162 & 0.839 & 0.271 & 0.271 & 13/50 \\
             & C2 & 0.112 & 0.946 & 0.198 & 0.201 & 10/50 \\
             & C3 & \sig{0.264}~/~0.255 & 0.333 & 0.674 & 0.433 & 48/50 \\
             & C4 & 0.235/0.220/0.135 & 0.448 & 0.585 & 0.433 & 38/50 \\
DeepSeek     & C1 & 0.109 & 0.956 & 0.184 & 0.183 & \sig{9/50} \\
V4.1-Flash   & C2 & \sig{0.228} & 0.737 & 0.344 & 0.341 & 16/50 \\
             & C3 & 0.203~/~0.207 & 0.347 & 0.660 & 0.391 & 49/50 \\
             & C4 & 0.218/0.178/0.119 & 0.399 & 0.620 & 0.429 & 41/50 \\
Claude       & C1 & 0.151 & 0.796 & 0.288 & 0.236 & 17/50 \\
Sonnet-5     & C2 & \sig{0.220} & 0.705 & 0.364 & 0.339 & 19/50 \\
             & C3 & 0.192~/~0.207 & 0.499 & 0.535 & 0.384 & 36/50 \\
             & C4 & 0.179/0.158/0.152 & 0.353 & 0.659 & 0.400 & 46/50 \\
\bottomrule
\end{tabular}
\end{table}

\subsection{Positive control}\label{sec:control}

C2 significantly beats C1 in five of six model--corpus pairs (Fig.~\ref{fig:deltas}, black),
so our setup can detect differences when they exist; the exception is the 9B model on
DepSign, where chain-of-thought is significantly \emph{worse} ($-0.050$). Asked directly, all
models over-predict severity: they rate 47--77\% of posts \emph{above} their gold label.
Chain-of-thought cuts this share sharply for the frontier models (to 18--24\% on
DepSeverity and 54\% on DepSign), hardly changes it for the 9B model on DepSeverity
(from 47\% to 45\%) and raises it on DepSign (from 67\% to 77\%).

The gain follows the drop in over-prediction, not model size. DeepSeek-V4.1-Flash
over-predicts most and gains most; the 9B model, whose over-prediction does not fall, gains
least and loses on DepSign.

\subsection{The unfair comparison}\label{sec:artifactsize}

\begin{figure}[!tb]
\centering
\includegraphics[width=\columnwidth]{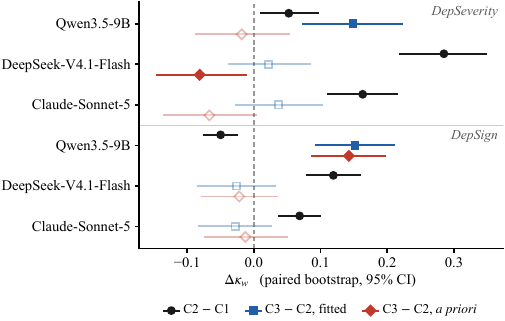}
\caption{Paired bootstrap $\Delta\kappa_w$ on the same posts (4000 resamples); solid markers
exclude zero at 95\%, faded hollow ones do not. C2 $-$ C1 uses no thresholds. C4 is in Table~\ref{tab:main}.}
\label{fig:deltas}
\end{figure}

Table~\ref{tab:main} gives all results. With fitted thresholds, C3 scores above C1 and C2 in
four of six model--corpus pairs and significantly beats C2 for the 9B model on both corpora
($+0.149$ and $+0.152$). With \emph{a priori} thresholds
(Fig.~\ref{fig:deltas}), DeepSeek-V4.1-Flash on DepSeverity shows a \emph{loss} of $-0.081$
(significant before correction). Across the nine \emph{a priori} comparisons with C2 for the frontier models and
for the 9B model's C3 on DepSeverity (five for C3, four for C4 under the same DSM-5 rule;
Table~\ref{tab:main}),
\textbf{none shows a significant $\kappa_w$ advantage for criteria extraction}; two ($-0.081$
for C3, $-0.062$ for C4) are significantly worse before correction, and another ($-0.050$, C4)
only barely so, its interval ending at $-0.001$. All three lose significance once the intervals
allow for run-to-run variation in the models' outputs (\S\ref{sec:stab}). All \emph{a priori}
gains come from the 9B model. On DepSign it keeps nearly all of its fitted gain ($+0.143$
against $+0.152$), and C4 under the same rule also beats C2 there ($+0.108$, significant before
correction); with C4 it also gains on DepSeverity ($+0.130$; \S\ref{sec:empty}).

On DepSeverity, the three fitted thresholds add $+0.167$ (Qwen3.5-9B), $+0.103$
(DeepSeek-V4.1-Flash) and $+0.104$ (Claude-Sonnet-5) in $\kappa_w$ over the \emph{a priori}
ones; on DepSign they add almost nothing ($+0.009$, $-0.004$, $-0.015$). For the 9B
model, the fitted thresholds become $[0.5, 1, 1.5]$, meaning ``no symptoms $\to$
\textsc{minimum}, any symptom $\to$ higher'', for a model that finds on average 0.26 of 9
symptoms. The thresholds do the work.

\textbf{The 9B exception is consistent with calibration.} Asked directly or with
chain-of-thought, the 9B model labels 444 and 526 of 706 DepSign posts as \textsc{severe} (50 in
gold). The \emph{a priori} rule, which needs five symptoms, labels 38 posts \textsc{severe}
and also significantly beats direct prediction ($+0.093$). The rule reduces over-prediction of
\textsc{severe}, but it still misses 47 of 50
\textsc{severe} posts (\S\ref{sec:safety}). The effect survives Holm correction and removing
the \texttt{[removed]} posts (\S\ref{sec:empty}), but after we recalibrate chain-of-thought with the
same 600 labels, no significant gap remains (\S\ref{sec:cal}).

\subsection{Giving C1 and C2 the same supervision}\label{sec:cal}

The stronger fairness test gives C1 and C2 the same access to labels that C3's fitted
thresholds get: we learn a monotone relabeling of
each condition's predicted label on the same 600-post split and freeze it. Such a map has
$\binom{2k-1}{k}$ forms on $k$ labels (35 for $k{=}4$, 10 for $k{=}3$), fewer than C3's
threshold options, so the test favors C3.

First, \textbf{the frontier models' chain-of-thought barely improves}: its best map leaves
the labels unchanged in three of four cases, and C3 with fitted thresholds still does not
significantly beat it in any case. These predictions are
already largely calibrated: C2 does not need the labels, but C3 does.

Second, \textbf{the 9B model on DepSeverity is the exception}. Its miscalibrated C2 improves
(from 0.340 to 0.379), which removes a quarter of the $+0.149$ gap, but $+0.109$ remains
(significant before correction). So for this model the unequal labels do not explain the
whole effect. C3's count runs from 0 to 9 while C2 gives one of four labels, so fitted thresholds have
more room than the relabeling. Capping C3's count at four levels (0, 1, 2, 3+) and searching
the same 35 maps leaves the $+0.109$ in place: the remaining gain comes from the extraction
itself, not from C3's finer score. Without labels, the result
depends on the questionnaire. With \emph{a priori} thresholds C3 falls to 0.322, below
recalibrated C2, so PHQ-9 extraction does not win; C4 under the same rule reaches 0.470,
$+0.091$ above recalibrated C2, just short of significance ($[-0.001, +0.181]$;
\appref{app:cal}).

Third, \textbf{C1 improves substantially but, on DepSeverity, still trails chain-of-thought}:
recalibration raises it from 0.219 to 0.330 on DepSeverity and from 0.109 to 0.246 on DepSign for DeepSeek-V4.1-Flash, so
its weakness is calibration. On DepSeverity, chain-of-thought with
\emph{no} labels (0.504) beats direct prediction with all 600 labels (0.330): reasoning is worth
more than the labels, and criteria extraction adds nothing measurable on top.

Fourth, \textbf{on DepSign the 9B model's gap is calibration}. Its chain-of-thought gains
more from labels than chain-of-thought does in any other case (from 0.112 to 0.219), through a
map that never predicts \textsc{severe}, and C3's lead with fitted thresholds shrinks to
a non-significant $+0.045$. Recalibrated direct prediction gains less (from 0.162 to 0.185) and
still trails C3 by $+0.078$ (significant before correction).
So labels close the gap through chain-of-thought but not through direct prediction.

\subsection{Other explanations}\label{sec:empty}

DepSeverity posts might show too few symptoms to extract, given their origin. On DepSign,
however, the frontier models extract about five times more symptoms (mean 2.31 and 2.29
against 0.45 and 0.46) and the 9B model seven times more (1.86 against 0.26), \textbf{and the
frontier models still show no advantage}. Extraction is faithful: 97.7--100\% of the frontier models'
quotes appear word for word in the post, and 92.4\% and 80.9\% of the 9B model's (90.4\% and 81.9\% in
its C4 runs).

Excluding the 53 \texttt{[removed]} test posts lowers every DepSign score, C3 most (C1 by
0.019--0.032, C2 by 0.026--0.045, C3 by 0.053--0.069). The frontier models' gap between C3 with fitted thresholds
and C2 then widens from about $-0.03$ to $-0.051$ for both, still not significant, and the 9B
model's \emph{a priori} gain remains significant ($+0.115$).

\textbf{A longer questionnaire helps, but beats C2 only for the 9B model.} BDI-II's 21 items
raise the mean count per post, from 0.26--0.46 to 0.72--1.25 on DepSeverity and from
1.86--2.31 to 3.31--4.22 on DepSign, where the maximum rises from 8 to 15. For the frontier
models on DepSeverity, C4 under the DSM-5 rule improves on C3 (0.479 and 0.466 against 0.423
and 0.395) but only reaches C2, and on neither corpus does C4 beat C2. For the 9B model it
does: under the DSM-5 rule, C4 beats C2 by $+0.130$ on DepSeverity and $+0.108$ on DepSign,
and C1 by $+0.182$ and $+0.058$, all significant before correction. On DepSeverity this
reverses the C3 result, where the same rule stays below C2. One reason is scale: the DSM-5
thresholds were set for nine criteria, and a 21-item count reaches them more easily, so they
sit close to what fitting would select. Fitting adds only $+0.025$ to the 9B model's C4, against
$+0.167$ to its C3 (\appref{app:cutoffs}). Under the second \emph{a priori} rule, BDI-II's own bands
rescaled to our count, C4 is significantly \emph{worse} than C2 for both frontier models on
both corpora and for the 9B model on DepSeverity (down to $-0.425$; Table~\ref{tab:main}),
because binary extraction finds too few items to reach them; for the 9B model on DepSign the
difference is $+0.023$, not significant.

\textbf{Prompt wording does not explain the result.} We reran C3 on DepSeverity with
DeepSeek-V4.1-Flash using two other prompts (2{,}612 calls; Table~\ref{tab:ablation}). A
\emph{permissive} prompt (a lower bar for \texttt{present}) finds 60\% more symptoms (from 0.45
to 0.72 per post) but scores slightly worse (fitted $\kappa_w$ from 0.526 to 0.507). A
\emph{symmetric} prompt (quotes also required for \texttt{absent}) scores $0.555$ but still does
not beat C2 with \emph{a priori} thresholds ($-0.071$, not significant), and raises the use of
\texttt{absent} only from 0.30\% to 0.35\%. Neither beats chain-of-thought under either rule, so
the rare use of \texttt{absent} (\S\ref{sec:agg}) comes from the posts, not our wording.

A better aggregation rule does not help: an ordinal model learned on the same nine criterion
labels, with far more free parameters, reaches only $0.156$ and $0.166$, significantly below
the count with fitted thresholds. Weighting its classes by inverse frequency raises it to 0.421--0.454 on
DepSeverity, still below the count with fitted thresholds, but it then misses only 14--26 of
56 \textsc{severe} posts, against 43--49 for the count (\appref{app:recall}).

\subsection{Separating structure, format and labeler (C2S)}\label{sec:c2s}

For the frontier models, \textbf{reasoning over the criteria changes nothing}: C2S $-$ C2
ranges from $-0.019$ to $+0.025$, none significant. \textbf{Nor does the output format}:
counting the C2S status lines with C3's rule matches counting C3's JSON in all eight
comparisons (within $\pm0.030$, none significant). \textbf{What matters is who assigns the
label.} On the same status lines, the \emph{a priori} count is never significantly better than
the model's own label and is significantly worse once ($-0.084$, DeepSeek-V4.1-Flash on
DepSeverity); the fitted count only ties it.

The 9B model differs on both counts. Reasoning over the criteria helps it ($+0.081$ on
DepSeverity and $+0.038$ on DepSign, both significant), again through calibration: it labels
fewer posts \textsc{severe} (from 14\% to 6\% and from 75\% to 61\%). The format matters too, but
not consistently: its free-text status lines mark more criteria than its JSON (0.56 against
0.26 per post on DepSeverity), and counting them beats counting C3 on DepSeverity by $+0.103$ \emph{a priori}
(significant; $+0.019$ fitted) yet trails it on DepSign ($-0.053$ \emph{a priori}, $-0.064$ fitted, the latter
significant).

\begin{figure}[!tb]
\centering
\includegraphics[width=\columnwidth]{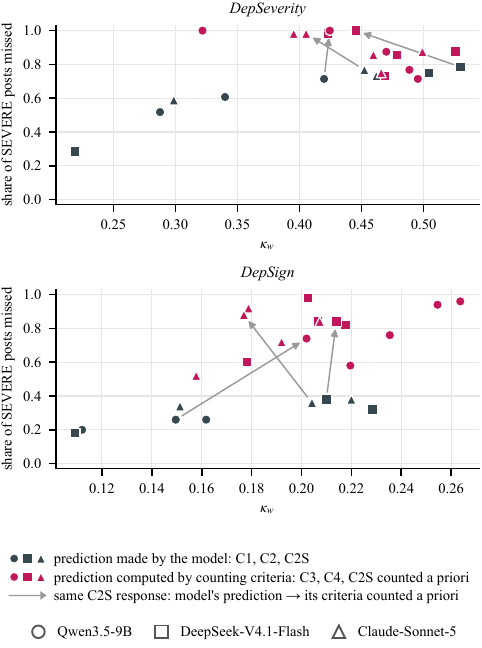}
\caption{Ordinal agreement against missed \textsc{severe} posts, one point per model and
condition (test sets). Each arrow follows one model's C2S responses from its own prediction to the \emph{a priori} count of
the criteria in the same responses.}
\label{fig:tradeoff}
\end{figure}

The ablation also locates the loss of \textsc{severe} posts (Fig.~\ref{fig:tradeoff}). With
the label from the model, C2S misses 40--44 of 56 \textsc{severe} posts on DepSeverity and
13--19 of 50 on DepSign, close to C2 (34--42 and 10--19). Counting the same status lines with
the \emph{a priori} rule misses 55--56 and 37--44. For every model, it is counting the criteria,
not reasoning about them, that loses severe cases. Length does not explain this either: C2S
writes 222--424 output tokens per post, fewer than C2 (363--614) for every model
(\appref{app:tokens}). Two 9B responses on DepSeverity end with ``\texttt{FINAL: unclear}'' and
are excluded from that cell ($n = 704$); all 7{,}836 C2S status lists parse. These comparisons
are exploratory (\appref{app:comp}).

\subsection{Ordinal agreement versus missed severe posts}\label{sec:safety}

\textbf{The last column of Table~\ref{tab:main} tells a different story.} On DepSign,
DeepSeek-V4.1-Flash under C1 has the lowest $\kappa_w$ in the table but misses only 9 of 50
\textsc{severe} posts; under C3 its $\kappa_w$ nearly doubles (from 0.109 to 0.203) but it misses 49 of 50. Across all
twelve comparisons, C1 to C2 and C2 to C3, the share of missed \textsc{severe} posts rises in
eleven, significantly in eight (95\% intervals, not corrected). The one decrease, not significant, is chain-of-thought for
Qwen3.5-9B on DepSign, which labels most posts as \textsc{severe}. Higher $\kappa_w$ often means more missed \textsc{severe} posts. Fig.~\ref{fig:tradeoff}
shows the pattern across conditions: predictions the model makes itself miss fewer
\textsc{severe} posts than predictions computed by counting criteria. BDI-II extraction misses fewer: under the DSM-5 rule on DepSign, C4 misses
30 and 26 of 50 \textsc{severe} posts against 42 for C3, but still more than
chain-of-thought (16 and 19), and its $\kappa_w$ is significantly lower
(\hyperref[app:comp]{Appendices~\ref*{app:comp}} and~\hyperref[app:recall]{\ref*{app:recall}}). For the 9B model, C4 misses 29 of 50
against 47 for C3 and 10 for chain-of-thought, which labels most posts as \textsc{severe}; on
DepSeverity it misses 49 of 56, against 56 for C3 and 34 for chain-of-thought.

Macro-F1, which ignores ordinal distance, mostly agrees with $\kappa_w$ (Table~\ref{tab:main}). The
exception is DepSign, where C3 beats C2 for the frontier models under both rules ($+0.044$ to
$+0.071$, before correction), entirely on the majority \textsc{moderate} class, which
chain-of-thought often raises to \textsc{severe}; C3 is worse on both smaller classes. We report $\kappa_w$ as the
main metric because the labels are ordinal and, unlike accuracy and MAE, $\kappa_w$ corrects
for chance: no LLM condition clearly beats majority-class accuracy (0.727 on DepSeverity).

\subsection{Run-to-run stability}\label{sec:stab}

Repeating C2 and C3 on DepSeverity changes $\kappa_w$ by $0.020$ and $0.008$ (Claude-Sonnet-5)
and $0.022$ and $0.020$ (DeepSeek-V4.1-Flash). Our bootstrap resamples posts but treats each model's answers as
fixed, so its intervals miss this run-to-run noise. We therefore widened each hosted model's intervals, treating its larger change as single-run noise (the
local model is deterministic). All six C2 $-$ C1 effects, the 9B model's gains and the fitted
margins over Dreaddit's features (\S\ref{sec:surface}) stay significant, but the three DSM-5-rule losses of
\S\ref{sec:artifactsize} ($-0.081$ for DeepSeek-V4.1-Flash on DepSeverity and both C4 losses on
DepSign) do not: a fresh run could erase them, so they do not show that extraction is worse. Structure is clearly better in one respect: between identical runs, C2 changes
\textbf{105} and \textbf{132} of 706 labels, and C3 only \textbf{21} and \textbf{29}.

\subsection{Can a human reader recover the labels?}\label{sec:human}

We do not know how reliable the gold labels are. As a test, the author, with over five years of
experience working with clinicians, annotated a stratified sample from each test split (100 and 102 posts) without seeing gold labels, model outputs or
any results, using a rubric based on BDI-II's bands and the nine DSM-5 criteria.

On DepSign, the annotator's $\kappa_w$ is $0.416$ $[0.248, 0.567]$, against 0.395
and 0.368 for the models' C2 on the same posts. The paired difference, $-0.021$
$[-0.174, +0.130]$, is too uncertain to show that human and model are equivalent, but
\textbf{the annotator's own interval is well above zero}: a reader following clinical criteria
recovers DepSign's labels, so its low $\kappa_w$ reflects a hard task with unbalanced classes,
not unrecoverable labels, and its results in \S\ref{sec:empty} are not noise.

\textbf{DepSeverity behaves differently.} The annotator reaches $0.199$ where the best model
condition reaches $0.485$, a paired gap of $+0.285$ $[+0.106, +0.450]$ in the \emph{model's}
favor, and reads 13 of 25 gold-\textsc{severe} posts as showing no depressive content. Models
can reproduce these labels, but a reader who follows clinical criteria cannot, which is hard to
explain if the labels follow clinical criteria and fits the idea that models learn patterns of
the annotation process. It also fits two other explanations, noise in the gold labels and a
mismatch between our rubric and the annotators' own; with one annotator and no access to the
original guidelines, we cannot tell these apart.

The difference does not come from the number of labels: collapsing DepSeverity's four labels to
three, as in DepSign, raises the annotator to at most $0.244$ and leaves the models unchanged, and the annotator's rank agreement
with gold, which does not depend on calibration, is $\rho = 0.251$ against $0.435$ on
DepSign.

\subsection{What predicts the labels}\label{sec:surface}

For DepSeverity, the recovered origin (\S\ref{sec:prov}) shows what the labels follow
instead. An ordinal
classifier~\cite{frank2001} fitted to Dreaddit-provided fields, without symptom annotations,
\textbf{recovers much of the label}: the source community alone reaches $\kappa_w = 0.196$, Dreaddit's
stress label $0.259$, and the fields that do not read the post (community, stress label, karma,
votes, comments) $0.333$. Adding text-derived features (length, Dreaddit's
sentiment score, LIWC word-category counts) gives $0.404$. An ordinal logistic model on TF-IDF features trained on the 2{,}824 labeled training posts
reaches $0.374$, above zero-shot direct prompting with every model (0.219--0.299), though it
sees labels that the prompts do not. The communities \textit{anxiety} and \textit{ptsd}, neither about depression,
provide 34 of the 56 gold-\textsc{severe} test posts.

On the same 706 posts, compared with the model using all of Dreaddit's features (0.404), \textbf{C3 with \emph{a priori}
thresholds is not significantly different for either model} ($+0.019$ for DeepSeek-V4.1-Flash,
$-0.009$ for Claude-Sonnet-5); against the non-text fields alone it is ahead by $+0.090$
(significant) and $+0.063$ (not). With fitted thresholds it is significantly ahead of the full
model ($+0.122$ and $+0.095$), which by our own standard (\S\ref{sec:cal}) is the fair
comparison. So Dreaddit's features are a lower bound, but a high one: without fitted
thresholds, criteria extraction is not significantly better than the full set, while
chain-of-thought scores above it without any labels.

This makes the result of \S\ref{sec:artifactsize} expected: the posts contain little
criterion-based signal, as the human test also showed. It is also a warning about the corpus
as a benchmark: a system can score well here without modeling depression severity.

\section{Limitations}

DepSeverity contains no depression community, which limits what it says about depression
severity; DepSign is drawn from depression communities but subsampled to 706 test posts, which
lowers power for its 50 \textsc{severe} posts. Our main tests are the eighteen comparisons in
Fig.~\ref{fig:deltas}, corrected together with Holm's method at $\alpha = 0.05$. The four
frontier C2 $-$ C1 effects, the 9B model's $+0.149$ on DepSeverity and all three of its DepSign
effects ($-0.050$, $+0.152$, $+0.143$) survive correction; its DepSeverity C2 $-$ C1 gain
($+0.052$) and the \emph{a priori} loss of $-0.081$ do not. For the frontier models we find no
significant advantage for criteria extraction, before or after correction. All other analyses
are exploratory and uncorrected: C4, C2S, comparisons of C3 with C1, the equal-supervision test, the
missed-\textsc{severe} counts, macro-F1, the Dreaddit-feature comparisons, the prompt variants and
the human probe. The conditions also differ in output format and length: C2 writes 363--614
output tokens per post, C3 152--251 and C4 371--579. C3 writes less than half as much as C2 and
C4 about as much, yet neither beats C2 for the frontier models, so length does not track the
ranking. The C2S ablation (\S\ref{sec:c2s}) controls format and content more directly; it
covers PHQ-9 only, and we did not vary the label wording or token budget further.

DepSign reports two annotators at $\kappa = 0.686$~\cite{sampath2022}, but DepSeverity's
annotation documents are not available, so we cannot measure its label noise; two of its posts
have identical text but different gold labels. Our human probe cannot settle this: it uses
one non-clinician annotator and our own rubric, so its DepSeverity result is a warning sign,
not evidence about the original annotators. We run each condition once, and the generation-variance check in
\S\ref{sec:stab} rests on one repeat per hosted model, on one corpus. All but seven of the
15{,}672 structured calls on the fit and test splits return the full inventory, and code
recovers the statuses from six malformed JSON responses; the seven exceptions are 9B C4
responses that omit one item (\texttt{sleep\_change}), which we count as not present (dropping
them changes no reported $\kappa_w$ by more than $0.001$). There is no exact \emph{a priori}
threshold for a 21-item count (\S\ref{sec:agg}): for the frontier models C4 fails with both
imperfect options and with fitted thresholds, but a better rule might do better. We score only whether each criterion or item
is present, while PHQ-9 scores each criterion 0--3 by frequency and BDI-II scores each item 0--3 by
severity. Binary scoring loses the intensity a post can express, which chain-of-thought can
still use. We did not test graded extraction; it could narrow the gap to chain-of-thought, and
because BDI-II's bands assume graded items, the C4 result with those bands (\S\ref{sec:empty})
may partly reflect our binary choice. Finally, because of compute limits we ran only one small
open-weights model; the calibration exception may not hold for other small models, and a
larger local model (for example 27B to 70B parameters) would show whether it disappears with
scale.

\section{Code, Data and Ethics}

The repository\footnote{\url{https://github.com/xinkaichen97/depseverity-artifact}}
contains the pipeline, exact prompts, result tables, worked examples and one cleaned record per
post per condition, from which every number except the \texttt{[removed]}-post analysis can be
recomputed without model calls. It contains \textbf{no post text}, since neither corpus has a
stated license.

This work uses public Reddit posts about depression, stress, abuse and self-harm. We did not
identify or contact authors or link people across datasets, we report only aggregate results,
and we share neither corpus. Nothing
here is fit for clinical use: the systems with the best ordinal agreement miss the most severe
cases (\S\ref{sec:safety}), so papers should report the direction of errors, not only their
size, for example per-class recall (\appref{app:recall}) or expected cost under
asymmetric error costs.

\section{Conclusion}

\textbf{For frontier models, criteria extraction shows no significant $\kappa_w$ advantage over
chain-of-thought across two corpora, two questionnaires and both threshold regimes.} Its apparent
gains depend on fitted thresholds, and its macro-F1 gains on DepSign come from the majority
class. The \emph{a priori} thresholds, equal-supervision test, BDI-II items, prompt variants,
learned aggregator, C2S ablation and human probe (\S\ref{sec:artifactsize}--\S\ref{sec:human})
do not rule out better extraction or aggregation, but they point to a mismatch between symptom
counts and the gold labels: only 3 of 56 gold-\textsc{severe} DepSeverity posts mention
self-harm, and the labels partly follow where posts were collected (\S\ref{sec:surface}).
Structure helps with \emph{a priori} thresholds only for the 9B model: on DepSign with either
questionnaire, consistent with correcting its over-prediction of \textsc{severe}, and on
DepSeverity with BDI-II's items, where the gain over recalibrated chain-of-thought falls just
short of significance. In every case the gains come with more missed severe cases, and holding
the criteria fixed shows that the counting step, not the reasoning over criteria, loses them.
Criteria extraction keeps one advantage: its outputs are far more stable across identical runs
than chain-of-thought's (\S\ref{sec:stab}) and can be audited criterion by criterion.

The lesson applies beyond mental health: \textbf{a pipeline with fitted thresholds and a
zero-shot baseline are not comparable systems}. Reporting the \emph{a priori} version next to
the fitted one costs nothing, because both use the same extractions, and reporting direct
prediction next to chain-of-thought is just as cheap: for the 9B model on DepSign, comparing only
against chain-of-thought would show $+0.143$, while direct prediction gives $+0.093$. Studies of
this kind should report both, and the direction of errors, not only agreement.

\onecolumn
\appendices

\section{All comparisons}\label{app:comp}

Tables~\ref{tab:allcomp} and~\ref{tab:allcomp-depsign} list every paired comparison between conditions. The eighteen marked
C are the main tests, corrected together (Limitations); the eight marked H survive Holm
correction. All other rows are exploratory, including every C4 comparison and every comparison of C3
with C1. Bold values have 95\% intervals that exclude
zero before any correction.

\begin{table}[!htbp]
\caption{All paired comparisons on DepSeverity ($\Delta\kappa_w$, 95\% paired bootstrap interval, 4000 resamples). Rows marked C are the main tests, corrected together; H = survives Holm correction at $\alpha=0.05$. C4 uses the DSM-5 rule for \emph{a priori}.}
\label{tab:allcomp}
\centering\footnotesize
\setlength{\tabcolsep}{6pt}
\begin{tabular}{lllc}
\toprule
\textbf{Model} & \textbf{Comparison} & $\Delta\kappa_w$ [95\% CI] & \\
\midrule
Qwen3.5-9B & C2 $-$ C1 & \textbf{+0.052} [+0.009, +0.098] & C \\
Qwen3.5-9B & C3 fitted $-$ C2 & \textbf{+0.149} [+0.072, +0.223] & C,H \\
Qwen3.5-9B & C3 \emph{a priori} $-$ C2 & $-$0.018 [$-$0.089, +0.054] & C \\
Qwen3.5-9B & C4 fitted $-$ C2 & \textbf{+0.155} [+0.082, +0.227] &  \\
Qwen3.5-9B & C4 \emph{a priori} $-$ C2 & \textbf{+0.130} [+0.050, +0.208] &  \\
Qwen3.5-9B & C3 fitted $-$ C1 & \textbf{+0.201} [+0.127, +0.276] &  \\
Qwen3.5-9B & C3 \emph{a priori} $-$ C1 & +0.034 [$-$0.037, +0.105] &  \\
Qwen3.5-9B & C4 fitted $-$ C1 & \textbf{+0.208} [+0.127, +0.282] &  \\
Qwen3.5-9B & C4 \emph{a priori} $-$ C1 & \textbf{+0.182} [+0.101, +0.259] &  \\
Qwen3.5-9B & C2S $-$ C2 & \textbf{+0.081} [+0.026, +0.133] &  \\
Qwen3.5-9B & C2S counted fitted $-$ C2S & \textbf{+0.087} [+0.039, +0.136] &  \\
Qwen3.5-9B & C2S counted \emph{a priori} $-$ C2S & +0.004 [$-$0.052, +0.059] &  \\
Qwen3.5-9B & C2S counted fitted $-$ C3 fitted & +0.019 [$-$0.043, +0.085] &  \\
Qwen3.5-9B & C2S counted \emph{a priori} $-$ C3 \emph{a priori} & \textbf{+0.103} [+0.053, +0.152] &  \\
DeepSeek-V4.1-Flash & C2 $-$ C1 & \textbf{+0.285} [+0.218, +0.349] & C,H \\
DeepSeek-V4.1-Flash & C3 fitted $-$ C2 & +0.022 [$-$0.039, +0.086] & C \\
DeepSeek-V4.1-Flash & C3 \emph{a priori} $-$ C2 & \textbf{$-$0.081} [$-$0.147, $-$0.010] & C \\
DeepSeek-V4.1-Flash & C4 fitted $-$ C2 & $-$0.035 [$-$0.107, +0.039] &  \\
DeepSeek-V4.1-Flash & C4 \emph{a priori} $-$ C2 & $-$0.026 [$-$0.100, +0.049] &  \\
DeepSeek-V4.1-Flash & C3 fitted $-$ C1 & \textbf{+0.307} [+0.237, +0.376] &  \\
DeepSeek-V4.1-Flash & C3 \emph{a priori} $-$ C1 & \textbf{+0.204} [+0.139, +0.270] &  \\
DeepSeek-V4.1-Flash & C4 fitted $-$ C1 & \textbf{+0.250} [+0.183, +0.312] &  \\
DeepSeek-V4.1-Flash & C4 \emph{a priori} $-$ C1 & \textbf{+0.259} [+0.192, +0.324] &  \\
DeepSeek-V4.1-Flash & C2S $-$ C2 & +0.025 [$-$0.018, +0.071] &  \\
DeepSeek-V4.1-Flash & C2S counted fitted $-$ C2S & +0.021 [$-$0.028, +0.071] &  \\
DeepSeek-V4.1-Flash & C2S counted \emph{a priori} $-$ C2S & \textbf{$-$0.084} [$-$0.139, $-$0.026] &  \\
DeepSeek-V4.1-Flash & C2S counted fitted $-$ C3 fitted & +0.024 [$-$0.018, +0.067] &  \\
DeepSeek-V4.1-Flash & C2S counted \emph{a priori} $-$ C3 \emph{a priori} & +0.022 [$-$0.016, +0.061] &  \\
Claude-Sonnet-5 & C2 $-$ C1 & \textbf{+0.163} [+0.110, +0.216] & C,H \\
Claude-Sonnet-5 & C3 fitted $-$ C2 & +0.037 [$-$0.029, +0.103] & C \\
Claude-Sonnet-5 & C3 \emph{a priori} $-$ C2 & $-$0.067 [$-$0.135, +0.005] & C \\
Claude-Sonnet-5 & C4 fitted $-$ C2 & $-$0.002 [$-$0.075, +0.072] &  \\
Claude-Sonnet-5 & C4 \emph{a priori} $-$ C2 & +0.004 [$-$0.068, +0.079] &  \\
Claude-Sonnet-5 & C3 fitted $-$ C1 & \textbf{+0.200} [+0.131, +0.271] &  \\
Claude-Sonnet-5 & C3 \emph{a priori} $-$ C1 & \textbf{+0.097} [+0.028, +0.167] &  \\
Claude-Sonnet-5 & C4 fitted $-$ C1 & \textbf{+0.161} [+0.096, +0.223] &  \\
Claude-Sonnet-5 & C4 \emph{a priori} $-$ C1 & \textbf{+0.167} [+0.102, +0.230] &  \\
Claude-Sonnet-5 & C2S $-$ C2 & $-$0.010 [$-$0.051, +0.032] &  \\
Claude-Sonnet-5 & C2S counted fitted $-$ C2S & +0.043 [$-$0.016, +0.099] &  \\
Claude-Sonnet-5 & C2S counted \emph{a priori} $-$ C2S & $-$0.047 [$-$0.107, +0.015] &  \\
Claude-Sonnet-5 & C2S counted fitted $-$ C3 fitted & $-$0.004 [$-$0.051, +0.043] &  \\
Claude-Sonnet-5 & C2S counted \emph{a priori} $-$ C3 \emph{a priori} & +0.010 [$-$0.029, +0.050] &  \\
\bottomrule
\end{tabular}
\end{table}

\begin{table}[!htbp]
\caption{All paired comparisons on DepSign, as in Table~\ref{tab:allcomp}.}
\label{tab:allcomp-depsign}
\centering\footnotesize
\setlength{\tabcolsep}{6pt}
\begin{tabular}{lllc}
\toprule
\textbf{Model} & \textbf{Comparison} & $\Delta\kappa_w$ [95\% CI] & \\
\midrule
Qwen3.5-9B & C2 $-$ C1 & \textbf{$-$0.050} [$-$0.077, $-$0.024] & C,H \\
Qwen3.5-9B & C3 fitted $-$ C2 & \textbf{+0.152} [+0.092, +0.212] & C,H \\
Qwen3.5-9B & C3 \emph{a priori} $-$ C2 & \textbf{+0.143} [+0.086, +0.198] & C,H \\
Qwen3.5-9B & C4 fitted $-$ C2 & \textbf{+0.123} [+0.064, +0.182] &  \\
Qwen3.5-9B & C4 \emph{a priori} $-$ C2 & \textbf{+0.108} [+0.055, +0.158] &  \\
Qwen3.5-9B & C3 fitted $-$ C1 & \textbf{+0.102} [+0.044, +0.162] &  \\
Qwen3.5-9B & C3 \emph{a priori} $-$ C1 & \textbf{+0.093} [+0.037, +0.148] &  \\
Qwen3.5-9B & C4 fitted $-$ C1 & \textbf{+0.074} [+0.017, +0.130] &  \\
Qwen3.5-9B & C4 \emph{a priori} $-$ C1 & \textbf{+0.058} [+0.008, +0.108] &  \\
Qwen3.5-9B & C2S $-$ C2 & \textbf{+0.038} [+0.008, +0.068] &  \\
Qwen3.5-9B & C2S counted fitted $-$ C2S & \textbf{+0.050} [+0.007, +0.093] &  \\
Qwen3.5-9B & C2S counted \emph{a priori} $-$ C2S & +0.052 [$-$0.002, +0.103] &  \\
Qwen3.5-9B & C2S counted fitted $-$ C3 fitted & \textbf{$-$0.064} [$-$0.129, $-$0.002] &  \\
Qwen3.5-9B & C2S counted \emph{a priori} $-$ C3 \emph{a priori} & $-$0.053 [$-$0.116, +0.010] &  \\
DeepSeek-V4.1-Flash & C2 $-$ C1 & \textbf{+0.119} [+0.078, +0.160] & C,H \\
DeepSeek-V4.1-Flash & C3 fitted $-$ C2 & $-$0.026 [$-$0.086, +0.033] & C \\
DeepSeek-V4.1-Flash & C3 \emph{a priori} $-$ C2 & $-$0.022 [$-$0.080, +0.036] & C \\
DeepSeek-V4.1-Flash & C4 fitted $-$ C2 & $-$0.011 [$-$0.074, +0.050] &  \\
DeepSeek-V4.1-Flash & C4 \emph{a priori} $-$ C2 & \textbf{$-$0.050} [$-$0.100, $-$0.001] &  \\
DeepSeek-V4.1-Flash & C3 fitted $-$ C1 & \textbf{+0.093} [+0.033, +0.152] &  \\
DeepSeek-V4.1-Flash & C3 \emph{a priori} $-$ C1 & \textbf{+0.097} [+0.038, +0.158] &  \\
DeepSeek-V4.1-Flash & C4 fitted $-$ C1 & \textbf{+0.108} [+0.047, +0.168] &  \\
DeepSeek-V4.1-Flash & C4 \emph{a priori} $-$ C1 & \textbf{+0.069} [+0.019, +0.118] &  \\
DeepSeek-V4.1-Flash & C2S $-$ C2 & $-$0.019 [$-$0.051, +0.013] &  \\
DeepSeek-V4.1-Flash & C2S counted fitted $-$ C2S & +0.004 [$-$0.046, +0.053] &  \\
DeepSeek-V4.1-Flash & C2S counted \emph{a priori} $-$ C2S & +0.004 [$-$0.046, +0.053] &  \\
DeepSeek-V4.1-Flash & C2S counted fitted $-$ C3 fitted & +0.011 [$-$0.042, +0.062] &  \\
DeepSeek-V4.1-Flash & C2S counted \emph{a priori} $-$ C3 \emph{a priori} & +0.007 [$-$0.037, +0.050] &  \\
Claude-Sonnet-5 & C2 $-$ C1 & \textbf{+0.069} [+0.036, +0.101] & C,H \\
Claude-Sonnet-5 & C3 fitted $-$ C2 & $-$0.028 [$-$0.083, +0.028] & C \\
Claude-Sonnet-5 & C3 \emph{a priori} $-$ C2 & $-$0.013 [$-$0.074, +0.052] & C \\
Claude-Sonnet-5 & C4 fitted $-$ C2 & $-$0.041 [$-$0.110, +0.024] &  \\
Claude-Sonnet-5 & C4 \emph{a priori} $-$ C2 & \textbf{$-$0.062} [$-$0.113, $-$0.011] &  \\
Claude-Sonnet-5 & C3 fitted $-$ C1 & +0.041 [$-$0.016, +0.097] &  \\
Claude-Sonnet-5 & C3 \emph{a priori} $-$ C1 & +0.056 [$-$0.009, +0.121] &  \\
Claude-Sonnet-5 & C4 fitted $-$ C1 & +0.027 [$-$0.036, +0.090] &  \\
Claude-Sonnet-5 & C4 \emph{a priori} $-$ C1 & +0.006 [$-$0.040, +0.055] &  \\
Claude-Sonnet-5 & C2S $-$ C2 & $-$0.016 [$-$0.039, +0.007] &  \\
Claude-Sonnet-5 & C2S counted fitted $-$ C2S & $-$0.026 [$-$0.064, +0.011] &  \\
Claude-Sonnet-5 & C2S counted \emph{a priori} $-$ C2S & $-$0.027 [$-$0.084, +0.029] &  \\
Claude-Sonnet-5 & C2S counted fitted $-$ C3 fitted & $-$0.014 [$-$0.058, +0.028] &  \\
Claude-Sonnet-5 & C2S counted \emph{a priori} $-$ C3 \emph{a priori} & $-$0.030 [$-$0.079, +0.015] &  \\
\bottomrule
\end{tabular}
\end{table}

\section{Thresholds}\label{app:cutoffs}

Table~\ref{tab:cutoffs} gives the thresholds behind every C3 and C4 cell. Two patterns stand out.
On DepSeverity, C3's fitted rules move the top threshold down to 1.5--2.5, so a post needs
only two or three marked criteria to be labeled \textsc{severe}; the DSM-5 rule needs five,
which few posts reach (Fig.~\ref{fig:counts}). On DepSign, the fitted C4 rules move the
top threshold \emph{up}, to 6.5 (9B), 8.5 and 10.5 of 21 items, so fitted C4 catches fewer
\textsc{severe} posts there than the DSM-5 rule does (\appref{app:recall}).

\begin{table}[!htbp]
\caption{Thresholds on the count of \texttt{present} criteria (C3: 0--9, C4: 0--21). Fitted thresholds maximize $\kappa_w$ on the 600-post fitting split.}
\label{tab:cutoffs}
\centering\footnotesize
\setlength{\tabcolsep}{3pt}
\begin{tabular}{lllll}
\toprule
\textbf{Corpus} & \textbf{Model} & \textbf{Cond.} & \textbf{fitted} & \textbf{\emph{a priori}} \\
\midrule
DepSeverity & Qwen3.5-9B & C3 & [0.5, 1, 1.5] & [0.5, 2.5, 4.5] \\
DepSeverity & Qwen3.5-9B & C4 & [0.5, 1.5, 3.5] & [0.5, 2.5, 4.5] \\
DepSeverity & DeepSeek-V4.1-Flash & C3 & [0.5, 1.5, 2.5] & [0.5, 2.5, 4.5] \\
DepSeverity & DeepSeek-V4.1-Flash & C4 & [0.5, 1.5, 3.5] & [0.5, 2.5, 4.5] \\
DepSeverity & Claude-Sonnet-5 & C3 & [0.5, 1.5, 2.5] & [0.5, 2.5, 4.5] \\
DepSeverity & Claude-Sonnet-5 & C4 & [0.5, 2.5, 5.5] & [0.5, 2.5, 4.5] \\
DepSign & Qwen3.5-9B & C3 & [0.5, 5.5] & [0.5, 4.5] \\
DepSign & Qwen3.5-9B & C4 & [0.5, 6.5] & [0.5, 4.5] \\
DepSign & DeepSeek-V4.1-Flash & C3 & [0.5, 5.5] & [0.5, 4.5] \\
DepSign & DeepSeek-V4.1-Flash & C4 & [0.5, 8.5] & [0.5, 4.5] \\
DepSign & Claude-Sonnet-5 & C3 & [0.5, 3.5] & [0.5, 4.5] \\
DepSign & Claude-Sonnet-5 & C4 & [0.5, 10.5] & [0.5, 4.5] \\
\bottomrule
\end{tabular}
\end{table}

\section{Giving C1 and C2 the same supervision}\label{app:cal}

Table~\ref{tab:cal} expands \S\ref{sec:cal}. Its relabeling column lists the predicted labels that the monotone map selected on
the fitting split changes; ``unchanged'' means no relabeling improved on the condition's own labels.
Table~\ref{tab:cal2} applies the same test to the \emph{a priori} rules and to C4.

\begin{table}[!htbp]
\caption{C1 and C2 given the same supervision as C3: a monotone relabeling of each condition's predicted label, fitted on the 600-post fitting split and frozen. The relabeling column lists the predicted labels it changes. The last column is C3 fitted minus the recalibrated condition, $\Delta\kappa_w$ with 95\% paired bootstrap interval; exploratory; bold intervals exclude zero.}
\label{tab:cal}
\centering\footnotesize
\setlength{\tabcolsep}{2.5pt}
\begin{tabular}{lllcccl}
\toprule
\textbf{Corpus} & \textbf{Model} & & raw & +cal & relabeling & C3 fitted $-$ +cal \\
\midrule
DepSeverity & Qwen3.5-9B & C1 & 0.288 & 0.320 & \textsc{mild}$\to$\textsc{min.}, \textsc{mod.}$\to$\textsc{mild}, \textsc{sev.}$\to$\textsc{mod.} & \textbf{+0.169} [+0.090, +0.246] \\
DepSeverity & Qwen3.5-9B & C2 & 0.340 & 0.379 & \textsc{mild}$\to$\textsc{min.}, \textsc{mod.}$\to$\textsc{mild} & \textbf{+0.109} [+0.027, +0.189] \\
DepSeverity & DeepSeek-V4.1-Flash & C1 & 0.219 & 0.330 & \textsc{mild}$\to$\textsc{min.}, \textsc{mod.}$\to$\textsc{min.}, \textsc{sev.}$\to$\textsc{mod.} & \textbf{+0.196} [+0.113, +0.277] \\
DepSeverity & DeepSeek-V4.1-Flash & C2 & 0.504 & 0.504 & unchanged & +0.022 [$-$0.039, +0.086] \\
DepSeverity & Claude-Sonnet-5 & C1 & 0.299 & 0.352 & \textsc{mild}$\to$\textsc{min.}, \textsc{mod.}$\to$\textsc{mild} & \textbf{+0.147} [+0.067, +0.226] \\
DepSeverity & Claude-Sonnet-5 & C2 & 0.462 & 0.462 & unchanged & +0.037 [$-$0.029, +0.103] \\
DepSign & Qwen3.5-9B & C1 & 0.162 & 0.185 & \textsc{sev.}$\to$\textsc{mod.} & \textbf{+0.078} [+0.011, +0.148] \\
DepSign & Qwen3.5-9B & C2 & 0.112 & 0.219 & \textsc{mod.}$\to$\textsc{not dep.}, \textsc{sev.}$\to$\textsc{mod.} & +0.045 [$-$0.024, +0.112] \\
DepSign & DeepSeek-V4.1-Flash & C1 & 0.109 & 0.246 & \textsc{mod.}$\to$\textsc{not dep.}, \textsc{sev.}$\to$\textsc{mod.} & $-$0.043 [$-$0.116, +0.031] \\
DepSign & DeepSeek-V4.1-Flash & C2 & 0.228 & 0.262 & \textsc{sev.}$\to$\textsc{mod.} & $-$0.059 [$-$0.119, +0.003] \\
DepSign & Claude-Sonnet-5 & C1 & 0.151 & 0.234 & \textsc{mod.}$\to$\textsc{not dep.}, \textsc{sev.}$\to$\textsc{mod.} & $-$0.042 [$-$0.112, +0.028] \\
DepSign & Claude-Sonnet-5 & C2 & 0.220 & 0.220 & unchanged & $-$0.028 [$-$0.083, +0.028] \\
\bottomrule
\end{tabular}
\end{table}

\begin{table}[!htbp]
\caption{The \emph{a priori} rules and C4 against C1 and C2. Every C1 and C2 prediction here is recalibrated on the 600-post fitting split with the monotone relabeling of Table~\ref{tab:cal}. $\Delta\kappa_w$ with 95\% paired bootstrap interval; exploratory; bold intervals exclude zero.}
\label{tab:cal2}
\centering\footnotesize
\setlength{\tabcolsep}{2.5pt}
\begin{tabular}{lllc}
\toprule
\textbf{Corpus} & \textbf{Model} & \textbf{Comparison} & $\Delta\kappa_w$ [95\% CI] \\
\midrule
DepSeverity & Qwen3.5-9B & C3 \emph{a priori} $-$ C2 & $-$0.058 [$-$0.136, +0.023] \\
DepSeverity & Qwen3.5-9B & C4 fitted $-$ C2 & \textbf{+0.116} [+0.024, +0.202] \\
DepSeverity & Qwen3.5-9B & C4 \emph{a priori} $-$ C2 & +0.091 [$-$0.001, +0.181] \\
DepSeverity & Qwen3.5-9B & C3 \emph{a priori} $-$ C1 & +0.002 [$-$0.072, +0.079] \\
DepSeverity & Qwen3.5-9B & C4 fitted $-$ C1 & \textbf{+0.176} [+0.085, +0.259] \\
DepSeverity & Qwen3.5-9B & C4 \emph{a priori} $-$ C1 & \textbf{+0.150} [+0.060, +0.235] \\
DepSeverity & DeepSeek-V4.1-Flash & C3 \emph{a priori} $-$ C2 & \textbf{$-$0.081} [$-$0.147, $-$0.010] \\
DepSeverity & DeepSeek-V4.1-Flash & C4 fitted $-$ C2 & $-$0.035 [$-$0.107, +0.039] \\
DepSeverity & DeepSeek-V4.1-Flash & C4 \emph{a priori} $-$ C2 & $-$0.026 [$-$0.100, +0.049] \\
DepSeverity & DeepSeek-V4.1-Flash & C3 \emph{a priori} $-$ C1 & \textbf{+0.094} [+0.013, +0.173] \\
DepSeverity & DeepSeek-V4.1-Flash & C4 fitted $-$ C1 & \textbf{+0.139} [+0.050, +0.226] \\
DepSeverity & DeepSeek-V4.1-Flash & C4 \emph{a priori} $-$ C1 & \textbf{+0.149} [+0.061, +0.236] \\
DepSeverity & Claude-Sonnet-5 & C3 \emph{a priori} $-$ C2 & $-$0.067 [$-$0.135, +0.005] \\
DepSeverity & Claude-Sonnet-5 & C4 fitted $-$ C2 & $-$0.002 [$-$0.075, +0.072] \\
DepSeverity & Claude-Sonnet-5 & C4 \emph{a priori} $-$ C2 & +0.004 [$-$0.068, +0.079] \\
DepSeverity & Claude-Sonnet-5 & C3 \emph{a priori} $-$ C1 & +0.043 [$-$0.039, +0.125] \\
DepSeverity & Claude-Sonnet-5 & C4 fitted $-$ C1 & \textbf{+0.107} [+0.023, +0.190] \\
DepSeverity & Claude-Sonnet-5 & C4 \emph{a priori} $-$ C1 & \textbf{+0.114} [+0.030, +0.199] \\
DepSign & Qwen3.5-9B & C3 \emph{a priori} $-$ C2 & +0.036 [$-$0.030, +0.101] \\
DepSign & Qwen3.5-9B & C4 fitted $-$ C2 & +0.016 [$-$0.049, +0.083] \\
DepSign & Qwen3.5-9B & C4 \emph{a priori} $-$ C2 & +0.001 [$-$0.061, +0.064] \\
DepSign & Qwen3.5-9B & C3 \emph{a priori} $-$ C1 & \textbf{+0.069} [+0.003, +0.138] \\
DepSign & Qwen3.5-9B & C4 fitted $-$ C1 & +0.050 [$-$0.019, +0.119] \\
DepSign & Qwen3.5-9B & C4 \emph{a priori} $-$ C1 & +0.034 [$-$0.031, +0.102] \\
DepSign & DeepSeek-V4.1-Flash & C3 \emph{a priori} $-$ C2 & $-$0.055 [$-$0.120, +0.010] \\
DepSign & DeepSeek-V4.1-Flash & C4 fitted $-$ C2 & $-$0.044 [$-$0.120, +0.027] \\
DepSign & DeepSeek-V4.1-Flash & C4 \emph{a priori} $-$ C2 & \textbf{$-$0.084} [$-$0.151, $-$0.018] \\
DepSign & DeepSeek-V4.1-Flash & C3 \emph{a priori} $-$ C1 & $-$0.039 [$-$0.112, +0.035] \\
DepSign & DeepSeek-V4.1-Flash & C4 fitted $-$ C1 & $-$0.028 [$-$0.098, +0.042] \\
DepSign & DeepSeek-V4.1-Flash & C4 \emph{a priori} $-$ C1 & \textbf{$-$0.068} [$-$0.130, $-$0.003] \\
DepSign & Claude-Sonnet-5 & C3 \emph{a priori} $-$ C2 & $-$0.013 [$-$0.074, +0.052] \\
DepSign & Claude-Sonnet-5 & C4 fitted $-$ C2 & $-$0.041 [$-$0.110, +0.024] \\
DepSign & Claude-Sonnet-5 & C4 \emph{a priori} $-$ C2 & \textbf{$-$0.062} [$-$0.113, $-$0.011] \\
DepSign & Claude-Sonnet-5 & C3 \emph{a priori} $-$ C1 & $-$0.027 [$-$0.106, +0.052] \\
DepSign & Claude-Sonnet-5 & C4 fitted $-$ C1 & $-$0.056 [$-$0.132, +0.021] \\
DepSign & Claude-Sonnet-5 & C4 \emph{a priori} $-$ C1 & \textbf{$-$0.077} [$-$0.139, $-$0.012] \\
\bottomrule
\end{tabular}
\end{table}

\section{Per-class recall}\label{app:recall}

Tables~\ref{tab:recall-depseverity} and~\ref{tab:recall-depsign} give recall for every class,
not only \textsc{severe}. The conditions with the highest $\kappa_w$ trade recall on the top
class for recall on the majority class: on DepSign, C3 with fitted thresholds raises
\textsc{moderate} recall to 0.68--0.88 and cuts \textsc{severe} recall to 0.02--0.28. The
learned aggregator of \S\ref{sec:empty} shows the same trade-off from the other side: with
inverse-frequency class weights it reaches $\kappa_w$ 0.421--0.454 on DepSeverity and misses
only 14--26 of 56 \textsc{severe} posts, against 43--49 for the count with fitted thresholds.

\begin{table}[!htbp]
\caption{Per-class recall on DepSeverity test ($n$ per class: 513, 58, 79, 56). Bold: highest recall per class within each model.}
\label{tab:recall-depseverity}
\centering\footnotesize
\setlength{\tabcolsep}{3pt}
\begin{tabular}{llcccc}
\toprule
\textbf{Model} & \textbf{Cond.} & \textsc{minimum} & \textsc{mild} & \textsc{moderate} & \textsc{severe} \\
\midrule
Qwen3.5-9B & C1 & 0.46 & 0.29 & \textbf{0.43} & \textbf{0.48} \\
 & C2 & 0.50 & 0.26 & 0.38 & 0.39 \\
 & C3 fitted & \textbf{0.92} & 0.00 & 0.35 & 0.23 \\
 & C3 \emph{a priori} & \textbf{0.92} & 0.36 & 0.03 & 0.00 \\
 & C4 fitted & 0.77 & 0.29 & 0.35 & 0.29 \\
 & C4 \emph{a priori} & 0.77 & 0.53 & 0.14 & 0.12 \\
 & C2S & 0.64 & 0.34 & 0.34 & 0.29 \\
 & C2S counted fitted & 0.83 & 0.31 & 0.19 & 0.27 \\
 & C2S counted \emph{a priori} & 0.83 & \textbf{0.62} & 0.11 & 0.00 \\
DeepSeek-V4.1-Flash & C1 & 0.32 & 0.12 & 0.29 & \textbf{0.71} \\
 & C2 & 0.80 & 0.36 & 0.23 & 0.25 \\
 & C3 fitted & \textbf{0.84} & 0.36 & 0.16 & 0.12 \\
 & C3 \emph{a priori} & \textbf{0.84} & 0.59 & 0.10 & 0.02 \\
 & C4 fitted & 0.70 & 0.34 & \textbf{0.30} & 0.27 \\
 & C4 \emph{a priori} & 0.70 & \textbf{0.67} & 0.16 & 0.14 \\
 & C2S & 0.83 & 0.50 & 0.24 & 0.21 \\
 & C2S counted fitted & 0.82 & 0.40 & 0.19 & 0.21 \\
 & C2S counted \emph{a priori} & 0.82 & 0.66 & 0.11 & 0.00 \\
Claude-Sonnet-5 & C1 & 0.33 & 0.36 & \textbf{0.47} & \textbf{0.41} \\
 & C2 & 0.72 & 0.34 & 0.24 & 0.27 \\
 & C3 fitted & \textbf{0.83} & 0.36 & 0.20 & 0.12 \\
 & C3 \emph{a priori} & \textbf{0.83} & 0.59 & 0.09 & 0.02 \\
 & C4 fitted & 0.58 & 0.52 & 0.32 & 0.14 \\
 & C4 \emph{a priori} & 0.58 & 0.52 & 0.28 & 0.25 \\
 & C2S & 0.80 & 0.38 & 0.16 & 0.23 \\
 & C2S counted fitted & 0.74 & 0.41 & 0.25 & 0.21 \\
 & C2S counted \emph{a priori} & 0.74 & \textbf{0.69} & 0.10 & 0.02 \\
\bottomrule
\end{tabular}
\end{table}

\begin{table}[!htbp]
\caption{Per-class recall on DepSign test ($n$ per class: 184, 472, 50). Bold: highest recall per class within each model.}
\label{tab:recall-depsign}
\centering\footnotesize
\setlength{\tabcolsep}{3pt}
\begin{tabular}{llccc}
\toprule
\textbf{Model} & \textbf{Cond.} & \textsc{not dep.} & \textsc{moderate} & \textsc{severe} \\
\midrule
Qwen3.5-9B & C1 & 0.22 & 0.24 & 0.74 \\
 & C2 & 0.13 & 0.16 & \textbf{0.80} \\
 & C3 fitted & \textbf{0.40} & \textbf{0.85} & 0.04 \\
 & C3 \emph{a priori} & \textbf{0.40} & 0.78 & 0.06 \\
 & C4 fitted & 0.38 & 0.70 & 0.24 \\
 & C4 \emph{a priori} & 0.38 & 0.52 & 0.42 \\
 & C2S & 0.21 & 0.25 & 0.74 \\
 & C2S counted fitted & 0.23 & 0.60 & 0.42 \\
 & C2S counted \emph{a priori} & 0.23 & 0.77 & 0.26 \\
DeepSeek-V4.1-Flash & C1 & 0.11 & 0.14 & \textbf{0.82} \\
 & C2 & \textbf{0.42} & 0.28 & 0.68 \\
 & C3 fitted & 0.28 & \textbf{0.88} & 0.02 \\
 & C3 \emph{a priori} & 0.28 & 0.79 & 0.16 \\
 & C4 fitted & 0.32 & 0.79 & 0.18 \\
 & C4 \emph{a priori} & 0.32 & 0.52 & 0.40 \\
 & C2S & 0.32 & 0.35 & 0.62 \\
 & C2S counted fitted & 0.26 & 0.81 & 0.16 \\
 & C2S counted \emph{a priori} & 0.26 & 0.81 & 0.16 \\
Claude-Sonnet-5 & C1 & 0.09 & 0.33 & \textbf{0.66} \\
 & C2 & \textbf{0.32} & 0.36 & 0.62 \\
 & C3 fitted & 0.24 & 0.68 & 0.28 \\
 & C3 \emph{a priori} & 0.24 & 0.82 & 0.16 \\
 & C4 fitted & 0.22 & \textbf{0.89} & 0.08 \\
 & C4 \emph{a priori} & 0.22 & 0.46 & 0.48 \\
 & C2S & 0.25 & 0.38 & 0.64 \\
 & C2S counted fitted & 0.21 & 0.42 & 0.56 \\
 & C2S counted \emph{a priori} & 0.21 & 0.84 & 0.12 \\
\bottomrule
\end{tabular}
\end{table}

\section{Output length}\label{app:tokens}

Table~\ref{tab:tokens} reports mean tokens per test call. C2 writes two to three times as
much as C3, since its prompt asks the model to discuss absent symptoms as well as present ones,
and C4 writes about as much as C2; C2S, which walks C2 through C3's criteria, writes less
than C2 for every model. The ranking of the conditions does not follow length.

\begin{table}[!htbp]
\caption{Mean tokens per test call, input / output, as reported by each provider.}
\label{tab:tokens}
\centering\footnotesize
\setlength{\tabcolsep}{3pt}
\begin{tabular}{llccccc}
\toprule
\textbf{Corpus} & \textbf{Model} & \textbf{C1} & \textbf{C2} & \textbf{C2S} & \textbf{C3} & \textbf{C4} \\
\midrule
DepSeverity & Qwen3.5-9B & 179 / 3 & 212 / 519 & 469 / 358 & 535 / 186 & 767 / 459 \\
DepSeverity & DeepSeek-V4.1-Flash & 162 / 2 & 194 / 363 & 438 / 222 & 499 / 152 & 712 / 371 \\
DepSeverity & Claude-Sonnet-5 & 238 / 5 & 282 / 545 & 642 / 382 & 737 / 227 & 1021 / 541 \\
DepSign & Qwen3.5-9B & 276 / 3 & 309 / 492 & 566 / 391 & 633 / 200 & 865 / 476 \\
DepSign & DeepSeek-V4.1-Flash & 262 / 2 & 294 / 429 & 538 / 242 & 600 / 171 & 813 / 404 \\
DepSign & Claude-Sonnet-5 & 369 / 4 & 413 / 614 & 773 / 424 & 871 / 251 & 1155 / 579 \\
\bottomrule
\end{tabular}
\end{table}

\section{Prompt variants}\label{app:ablation}

Table~\ref{tab:ablation} gives the C3 prompt variants of \S\ref{sec:empty} in full.

\begin{table}[!htbp]
\caption{C3 prompt variants, DeepSeek-V4.1-Flash on DepSeverity (same nine criteria and JSON schema; wording only). Last two columns: $\Delta\kappa_w$ against C2, with 95\% paired bootstrap interval; bold intervals exclude zero.}
\label{tab:ablation}
\centering\footnotesize
\setlength{\tabcolsep}{2pt}
\begin{tabular}{lcccccc}
\toprule
\textbf{Variant} & pres./post & \texttt{absent} & fitted & \emph{a pr.} & vs C2, fitted & vs C2, \emph{a pr.} \\
\midrule
original & 0.45 & 0.30\% & 0.526 & 0.423 & +0.022 [$-$0.039, +0.086] & \textbf{$-$0.081} [$-$0.147, $-$0.010] \\
permissive & 0.72 & 0.25\% & 0.507 & 0.425 & +0.002 [$-$0.059, +0.067] & \textbf{$-$0.079} [$-$0.145, $-$0.012] \\
symmetric evidence & 0.47 & 0.35\% & 0.555 & 0.433 & +0.051 [$-$0.011, +0.116] & $-$0.071 [$-$0.138, +0.000] \\
\bottomrule
\end{tabular}
\end{table}

\clearpage
\section{Prompts}\label{app:prompts}

The exact prompts, as sent. \texttt{\{text\}} is the post. For DepSign, the label list in C1
and C2 is \emph{not depression, moderate, severe}. C3P and C3S are the prompt variants of
\appref{app:ablation}; their user prompt is C3's. C2S is the ablation of \S\ref{sec:c2s}; its
user prompt is C2's.

\noindent\textbf{C1 (direct), system prompt}\par\nobreak\smallskip\nobreak
{\ttfamily\scriptsize\raggedright\noindent You are assessing the severity of depression expressed in a single social media post.\\
Reply with exactly one word, the severity level, and nothing else.\\
The permitted levels, from lowest to highest, are: minimum, mild, moderate, severe.\par}\medskip

\noindent\textbf{C1 (direct), user prompt}\par\nobreak\smallskip\nobreak
{\ttfamily\scriptsize\raggedright\noindent Post:\\
\textquotedbl{}\textquotedbl{}\textquotedbl{}\\
\{text\}\\
\textquotedbl{}\textquotedbl{}\textquotedbl{}\\
\mbox{}\\
Severity level:\par}\medskip

\noindent\textbf{C2 (chain-of-thought), system prompt}\par\nobreak\smallskip\nobreak
{\ttfamily\scriptsize\raggedright\noindent You are assessing the severity of depression expressed in a single social media post.\\
\mbox{}\\
First reason step by step about which depressive symptoms are and are not evident in the post, and how strongly each is expressed. Then state the overall severity level.\\
\mbox{}\\
The permitted levels, from lowest to highest, are: minimum, mild, moderate, severe.\\
\mbox{}\\
End your reply with a final line in exactly this form:\\
FINAL: \textless{}level\textgreater{}\par}\medskip

\noindent\textbf{C2 (chain-of-thought), user prompt}\par\nobreak\smallskip\nobreak
{\ttfamily\scriptsize\raggedright\noindent Post:\\
\textquotedbl{}\textquotedbl{}\textquotedbl{}\\
\{text\}\\
\textquotedbl{}\textquotedbl{}\textquotedbl{}\par}\medskip

\noindent\textbf{C2S (criteria-structured chain-of-thought), system prompt}\par\nobreak\smallskip\nobreak
{\ttfamily\scriptsize\raggedright\noindent You are assessing the severity of depression expressed in a single social media post.\\
\mbox{}\\
First go through the nine symptoms below in order, one line each, in exactly this form:\\
\textless{}number\textgreater{}. \textless{}symptom\textgreater{}: present\textbar{}absent\textbar{}unclear - \textless{}brief reason\textgreater{}\\
Write each symptom name exactly as given below. Use \textquotedbl{}present\textquotedbl{} when the post gives evidence that the writer experiences the symptom, \textquotedbl{}absent\textquotedbl{} when it gives evidence that the writer does not, and \textquotedbl{}unclear\textquotedbl{} when it says neither.\\
\mbox{}\\
The nine symptoms are:\\
  1. \textquotedbl{}anhedonia\textquotedbl{} - little interest or pleasure in doing things\\
  2. \textquotedbl{}depressed\_mood\textquotedbl{} - feeling down, depressed, or hopeless\\
  3. \textquotedbl{}sleep\textquotedbl{} - trouble falling or staying asleep, or sleeping too much\\
  4. \textquotedbl{}fatigue\textquotedbl{} - feeling tired or having little energy\\
  5. \textquotedbl{}appetite\textquotedbl{} - poor appetite, overeating, or weight change\\
  6. \textquotedbl{}worthlessness\textquotedbl{} - feeling bad about yourself, worthless, or excessively guilty\\
  7. \textquotedbl{}concentration\textquotedbl{} - trouble concentrating on things\\
  8. \textquotedbl{}psychomotor\textquotedbl{} - moving or speaking noticeably slowly, or being restless and fidgety\\
  9. \textquotedbl{}self\_harm\textquotedbl{} - thoughts that you would be better off dead, or of hurting yourself\\
\mbox{}\\
Then reason about how severe the depression expressed in the post is overall, and state the overall severity level.\\
\mbox{}\\
The permitted levels, from lowest to highest, are: minimum, mild, moderate, severe.\\
\mbox{}\\
End your reply with a final line in exactly this form:\\
FINAL: \textless{}level\textgreater{}\par}\medskip

\noindent\textbf{C3 (PHQ-9 extraction), system prompt}\par\nobreak\smallskip\nobreak
{\ttfamily\scriptsize\raggedright\noindent You are annotating a single social media post for the presence of nine specific symptoms. You are not rating, scoring, or diagnosing anything - you only report, for each symptom, whether the post gives evidence of it.\\
\mbox{}\\
The nine symptoms are:\\
  1. \textquotedbl{}anhedonia\textquotedbl{} - little interest or pleasure in doing things\\
  2. \textquotedbl{}depressed\_mood\textquotedbl{} - feeling down, depressed, or hopeless\\
  3. \textquotedbl{}sleep\textquotedbl{} - trouble falling or staying asleep, or sleeping too much\\
  4. \textquotedbl{}fatigue\textquotedbl{} - feeling tired or having little energy\\
  5. \textquotedbl{}appetite\textquotedbl{} - poor appetite, overeating, or weight change\\
  6. \textquotedbl{}worthlessness\textquotedbl{} - feeling bad about yourself, worthless, or excessively guilty\\
  7. \textquotedbl{}concentration\textquotedbl{} - trouble concentrating on things\\
  8. \textquotedbl{}psychomotor\textquotedbl{} - moving or speaking noticeably slowly, or being restless and fidgety\\
  9. \textquotedbl{}self\_harm\textquotedbl{} - thoughts that you would be better off dead, or of hurting yourself\\
\mbox{}\\
For each symptom return one of:\\
  \textquotedbl{}present\textquotedbl{} - the post gives positive evidence that the writer experiences it\\
  \textquotedbl{}absent\textquotedbl{}  - the post gives positive evidence that the writer does NOT experience it\\
  \textquotedbl{}unclear\textquotedbl{} - the post does not say either way\\
\mbox{}\\
Use \textquotedbl{}unclear\textquotedbl{} when the post is simply silent about a symptom. Do not use \textquotedbl{}absent\textquotedbl{} merely because a symptom is unmentioned.\\
\mbox{}\\
When and only when a symptom is \textquotedbl{}present\textquotedbl{}, also return \textquotedbl{}evidence\textquotedbl{}: a short span copied verbatim from the post. Copy it exactly; do not paraphrase.\\
\mbox{}\\
Reply with JSON only - no preamble, no code fence, no commentary. Schema:\\
\{\textquotedbl{}\textless{}symptom\textgreater{}\textquotedbl{}: \{\textquotedbl{}status\textquotedbl{}: \textquotedbl{}present\textbar{}absent\textbar{}unclear\textquotedbl{}, \textquotedbl{}evidence\textquotedbl{}: \textquotedbl{}\textless{}verbatim span or empty\textgreater{}\textquotedbl{}\}, ...\}\\
Include all nine symptom keys exactly as written above.\par}\medskip

\noindent\textbf{C3 (PHQ-9 extraction), user prompt}\par\nobreak\smallskip\nobreak
{\ttfamily\scriptsize\raggedright\noindent Post:\\
\textquotedbl{}\textquotedbl{}\textquotedbl{}\\
\{text\}\\
\textquotedbl{}\textquotedbl{}\textquotedbl{}\par}\medskip

\noindent\textbf{C4 (BDI-II extraction), system prompt}\par\nobreak\smallskip\nobreak
{\ttfamily\scriptsize\raggedright\noindent You are annotating a single social media post for the presence of twenty-one specific symptoms. You are not rating, scoring, or diagnosing anything - you only report, for each symptom, whether the post gives evidence of it.\\
\mbox{}\\
The twenty-one symptoms are:\\
  1. \textquotedbl{}sadness\textquotedbl{} - feeling sad or unhappy\\
  2. \textquotedbl{}pessimism\textquotedbl{} - feeling discouraged or hopeless about the future\\
  3. \textquotedbl{}past\_failure\textquotedbl{} - feeling like a failure, or dwelling on past failures\\
  4. \textquotedbl{}loss\_of\_pleasure\textquotedbl{} - getting less pleasure from things previously enjoyed\\
  5. \textquotedbl{}guilt\textquotedbl{} - feeling guilty\\
  6. \textquotedbl{}punishment\textquotedbl{} - feeling one is being punished, or deserves punishment\\
  7. \textquotedbl{}self\_dislike\textquotedbl{} - disliking oneself, or having lost confidence in oneself\\
  8. \textquotedbl{}self\_criticism\textquotedbl{} - blaming or criticising oneself\\
  9. \textquotedbl{}suicidal\textquotedbl{} - thoughts of killing oneself, or of being better off dead\\
  10. \textquotedbl{}crying\textquotedbl{} - crying, or being unable to cry when one wants to\\
  11. \textquotedbl{}agitation\textquotedbl{} - feeling restless, agitated, or keyed up\\
  12. \textquotedbl{}loss\_of\_interest\textquotedbl{} - having lost interest in other people or activities\\
  13. \textquotedbl{}indecisiveness\textquotedbl{} - finding it harder than usual to make decisions\\
  14. \textquotedbl{}worthlessness\textquotedbl{} - feeling worthless, or of no value\\
  15. \textquotedbl{}loss\_of\_energy\textquotedbl{} - having less energy than usual\\
  16. \textquotedbl{}sleep\_change\textquotedbl{} - sleeping more or less than usual, or broken sleep\\
  17. \textquotedbl{}irritability\textquotedbl{} - being more irritable than usual\\
  18. \textquotedbl{}appetite\_change\textquotedbl{} - eating more or less than usual, or appetite change\\
  19. \textquotedbl{}concentration\textquotedbl{} - finding it harder than usual to concentrate\\
  20. \textquotedbl{}fatigue\textquotedbl{} - being too tired to do many of the things one used to do\\
  21. \textquotedbl{}loss\_of\_interest\_sex\textquotedbl{} - reduced interest in sex\\
\mbox{}\\
For each symptom return one of:\\
  \textquotedbl{}present\textquotedbl{} - the post gives positive evidence that the writer experiences it\\
  \textquotedbl{}absent\textquotedbl{}  - the post gives positive evidence that the writer does NOT experience it\\
  \textquotedbl{}unclear\textquotedbl{} - the post does not say either way\\
\mbox{}\\
Use \textquotedbl{}unclear\textquotedbl{} when the post is simply silent about a symptom. Do not use \textquotedbl{}absent\textquotedbl{} merely because a symptom is unmentioned.\\
\mbox{}\\
When and only when a symptom is \textquotedbl{}present\textquotedbl{}, also return \textquotedbl{}evidence\textquotedbl{}: a short span copied verbatim from the post. Copy it exactly; do not paraphrase.\\
\mbox{}\\
Reply with JSON only - no preamble, no code fence, no commentary. Schema:\\
\{\textquotedbl{}\textless{}symptom\textgreater{}\textquotedbl{}: \{\textquotedbl{}status\textquotedbl{}: \textquotedbl{}present\textbar{}absent\textbar{}unclear\textquotedbl{}, \textquotedbl{}evidence\textquotedbl{}: \textquotedbl{}\textless{}verbatim span or empty\textgreater{}\textquotedbl{}\}, ...\}\\
Include all twenty-one symptom keys exactly as written above.\par}\medskip

\noindent\textbf{C3P (permissive variant), system prompt}\par\nobreak\smallskip\nobreak
{\ttfamily\scriptsize\raggedright\noindent You are annotating a single social media post for the presence of nine specific symptoms. You are not rating, scoring, or diagnosing anything - you only report, for each symptom, whether the post indicates it.\\
\mbox{}\\
The nine symptoms are:\\
  1. \textquotedbl{}anhedonia\textquotedbl{} - little interest or pleasure in doing things\\
  2. \textquotedbl{}depressed\_mood\textquotedbl{} - feeling down, depressed, or hopeless\\
  3. \textquotedbl{}sleep\textquotedbl{} - trouble falling or staying asleep, or sleeping too much\\
  4. \textquotedbl{}fatigue\textquotedbl{} - feeling tired or having little energy\\
  5. \textquotedbl{}appetite\textquotedbl{} - poor appetite, overeating, or weight change\\
  6. \textquotedbl{}worthlessness\textquotedbl{} - feeling bad about yourself, worthless, or excessively guilty\\
  7. \textquotedbl{}concentration\textquotedbl{} - trouble concentrating on things\\
  8. \textquotedbl{}psychomotor\textquotedbl{} - moving or speaking noticeably slowly, or being restless and fidgety\\
  9. \textquotedbl{}self\_harm\textquotedbl{} - thoughts that you would be better off dead, or of hurting yourself\\
\mbox{}\\
For each symptom return one of:\\
  \textquotedbl{}present\textquotedbl{} - the post indicates the writer experiences it. Count it as present if the writer states it directly, describes it in their own words, or describes circumstances or behaviour from which it reasonably follows. Do not require a clinical phrasing, and do not require the writer to name the symptom.\\
  \textquotedbl{}absent\textquotedbl{}  - the post indicates the writer does NOT experience it\\
  \textquotedbl{}unclear\textquotedbl{} - the post gives no indication either way\\
\mbox{}\\
Err toward \textquotedbl{}present\textquotedbl{} when a reading of the post supports it. Reserve \textquotedbl{}unclear\textquotedbl{} for symptoms the post genuinely does not touch on.\\
\mbox{}\\
When a symptom is \textquotedbl{}present\textquotedbl{}, also return \textquotedbl{}evidence\textquotedbl{}: a short span copied verbatim from the post that supports the judgement. Copy it exactly; do not paraphrase.\\
\mbox{}\\
Reply with JSON only - no preamble, no code fence, no commentary. Schema:\\
\{\textquotedbl{}\textless{}symptom\textgreater{}\textquotedbl{}: \{\textquotedbl{}status\textquotedbl{}: \textquotedbl{}present\textbar{}absent\textbar{}unclear\textquotedbl{}, \textquotedbl{}evidence\textquotedbl{}: \textquotedbl{}\textless{}verbatim span or empty\textgreater{}\textquotedbl{}\}, ...\}\\
Include all nine symptom keys exactly as written above.\par}\medskip

\noindent\textbf{C3S (symmetric variant), system prompt}\par\nobreak\smallskip\nobreak
{\ttfamily\scriptsize\raggedright\noindent You are annotating a single social media post for the presence of nine specific symptoms. You are not rating, scoring, or diagnosing anything - you only report, for each symptom, what the post says about it.\\
\mbox{}\\
The nine symptoms are:\\
  1. \textquotedbl{}anhedonia\textquotedbl{} - little interest or pleasure in doing things\\
  2. \textquotedbl{}depressed\_mood\textquotedbl{} - feeling down, depressed, or hopeless\\
  3. \textquotedbl{}sleep\textquotedbl{} - trouble falling or staying asleep, or sleeping too much\\
  4. \textquotedbl{}fatigue\textquotedbl{} - feeling tired or having little energy\\
  5. \textquotedbl{}appetite\textquotedbl{} - poor appetite, overeating, or weight change\\
  6. \textquotedbl{}worthlessness\textquotedbl{} - feeling bad about yourself, worthless, or excessively guilty\\
  7. \textquotedbl{}concentration\textquotedbl{} - trouble concentrating on things\\
  8. \textquotedbl{}psychomotor\textquotedbl{} - moving or speaking noticeably slowly, or being restless and fidgety\\
  9. \textquotedbl{}self\_harm\textquotedbl{} - thoughts that you would be better off dead, or of hurting yourself\\
\mbox{}\\
For each symptom return one of:\\
  \textquotedbl{}present\textquotedbl{} - the post gives evidence that the writer experiences it\\
  \textquotedbl{}absent\textquotedbl{}  - the post gives evidence that the writer does NOT experience it. This includes the writer denying the symptom, describing its opposite, or describing functioning that is incompatible with it.\\
  \textquotedbl{}unclear\textquotedbl{} - the post does not address the symptom\\
\mbox{}\\
For BOTH \textquotedbl{}present\textquotedbl{} and \textquotedbl{}absent\textquotedbl{} you must return \textquotedbl{}evidence\textquotedbl{}: a short span copied verbatim from the post that supports the judgement. If you cannot quote a supporting span, the correct answer is \textquotedbl{}unclear\textquotedbl{}. Copy spans exactly; do not paraphrase.\\
\mbox{}\\
Reply with JSON only - no preamble, no code fence, no commentary. Schema:\\
\{\textquotedbl{}\textless{}symptom\textgreater{}\textquotedbl{}: \{\textquotedbl{}status\textquotedbl{}: \textquotedbl{}present\textbar{}absent\textbar{}unclear\textquotedbl{}, \textquotedbl{}evidence\textquotedbl{}: \textquotedbl{}\textless{}verbatim span or empty\textgreater{}\textquotedbl{}\}, ...\}\\
Include all nine symptom keys exactly as written above.\par}\medskip

\end{document}